\documentclass{article} 
\usepackage{iclr2026_conference,times}

\usepackage{amsmath,amsfonts,bm}

\def\eqref#1{equation~\ref{#1}}

\def\1{\bm{1}}

\DeclareMathAlphabet{\mathsfit}{\encodingdefault}{\sfdefault}{m}{sl}
\SetMathAlphabet{\mathsfit}{bold}{\encodingdefault}{\sfdefault}{bx}{n}

\usepackage{hyperref}
\usepackage{url}

\usepackage{multirow}
\usepackage{booktabs}
\usepackage{graphicx}
\usepackage{svg}
\usepackage{microtype}
\usepackage{cleveref}
\usepackage{wrapfig}
\usepackage{graphicx}
\usepackage{subcaption}

\makeatletter
\@ifundefined{pdfobjcompresslevel}{}{%
  \pdfobjcompresslevel=0 
}
\makeatother

\expandafter\def\expandafter\normalsize\expandafter{%
    \normalsize%
    \setlength\abovedisplayskip{0pt}%
    \setlength\belowdisplayskip{8pt}%
    \setlength\abovedisplayshortskip{-8pt}%
    \setlength\belowdisplayshortskip{2pt}%
}

\iclrfinalcopy
\title{AdvEvo-MARL: Shaping Internalized Safety through Adversarial Co-Evolution in Multi-Agent Reinforcement Learning}

\author{
\centering
\begin{tabular}[t]{c}
Zhenyu Pan$^{1}$, Yiting Zhang$^{2}$, Zhuo Liu$^{3}$, Yolo Yunlong Tang$^{3}$, Zeliang Zhang$^{3}$, \\ Haozheng Luo$^{1}$,
Yuwei Han$^{2}$, Jianshu Zhang$^{1}$, Dennis Wu$^{1}$, Hong-Yu Chen$^{1}$, Haoran Lu$^{1}$, \\ Haoyang Fang$^{4}$,
Manling Li$^{1}$, Chenliang Xu$^{3}$, Philip S.~Yu$^{2}$, Han Liu$^{1}$\\[6pt]
$^{1}$\textbf{Northwestern University} \quad
$^{2}$\textbf{University of Illinois at Chicago} \\
$^{3}$\textbf{University of Rochester} \quad
$^{4}$\textbf{Carnegie Mellon University}
\end{tabular}
}

\begin{document}

\maketitle

\begin{abstract}

LLM-based multi-agent systems excel at planning, tool use, and role coordination, but their openness and interaction complexity also expose them to jailbreak, prompt-injection, and adversarial collaboration. Existing defenses fall into two lines: (i) self-verification that asks each agent to pre-filter unsafe instructions before execution, and (ii) external guard modules that police behaviors. The former often underperforms because a standalone agent lacks sufficient capacity to detect cross-agent unsafe chains and delegation-induced risks; the latter increases system overhead and creates a single-point-of-failure—once compromised, system-wide safety collapses, and adding more guards worsens cost and complexity. To solve these challenges, we propose AdvEvo-MARL, a co-evolutionary multi-agent reinforcement learning framework that internalizes safety into task agents. Rather than relying on external guards, AdvEvo-MARL jointly optimizes attackers (which synthesize evolving jailbreak prompts) and defenders (task agents trained to both accomplish their duties and resist attacks) in adversarial learning environments. To stabilize learning and foster cooperation, we introduce a public baseline for advantage estimation: agents within the same functional group share a group-level mean-return baseline, enabling lower-variance updates and stronger intra-group coordination. Across representative attack scenarios, AdvEvo-MARL consistently keeps attack-success rate (ASR) below 20\%, whereas baselines reach up to 38.33\%, while preserving—and sometimes improving—task accuracy (up to +3.67\% on reasoning tasks). These results show that safety and utility can be jointly improved without relying on extra guard agents or added system overhead.

\end{abstract}

\section{Introduction}
LLM-based agents exhibit advanced capabilities in software engineering \citep{pan2025code,pan2024codevbenchllmsunderstanddevelopercentric}, question answering system \citep{pan2024chain,pan2024convcoaimprovingopendomainquestion}, and scientific discovery \citep{shao2025sciscigpt}. Building on this progress, multi-agent systems (MAS) coordinate specialized agents with diverse expertise to harness collective intelligence for solving increasingly complex real-world problems. However, as MAS become more capable, they also face growing safety challenges \citep{raza2025trism}. On one hand, MAS inherit vulnerabilities from single agents, particularly their susceptibility to jailbreak attacks, where malicious actors attempt to bypass safety guardrails. On the other hand, the complex interaction dynamics among agents, along with the presence of potentially unauthorized or adversarial agents, significantly expand the attack surface beyond that of isolated systems \citep{he2025red}.

To mitigate these risks, researchers mainly explore two broad categories of defense: (i) empowering each agent to locally verify the benignness of its inputs before generating responses \citep{autoinject}, and (ii) deploying external inspector agents to monitor and regulate information flow throughout interactions \citep{xiang2025guardagent}. While these approaches are effective to some extent, they suffer from notable limitations. External guard agents introduce a single point of failure—once compromised, the system is left defenseless—and scaling up the number of guards quickly incurs prohibitive computational costs, rendering them impractical for large-scale deployments \citep{chennabasappa2025llamafirewall}. Meanwhile, individual agents have limited capacity to detect or resist sophisticated, cross-agent attacks, making self-verification in isolation insufficient \citep{zhu2025safescientist}. A natural intuition is to embed safety awareness within task agents through targeted safety training. Yet training agents individually overlooks the collaborative dynamics required for effective multi-agent defense, and conventional safety training based on static datasets often leads to overfitting and poor generalization against adaptive adversaries \citep{geissler2024concept}.

To address these challenges, we introduce AdvEvo-MARL, a co-evolutionary multi-agent reinforcement learning (MARL) framework that embeds safety awareness directly within task agents. The core idea is to jointly evolve attackers, which generate increasingly sophisticated jailbreak prompts, and defenders, which must both resist these attacks and fulfill their assigned tasks. AdvEvo-MARL initializes training with a curated pool of adversarial prompts derived from representative attack strategies. Since attackers lack prior knowledge of effective jailbreak tactics, we first warm them up using carefully designed seed prompts from this pool by supervised fine-tuning (SFT). During following MARL, attackers rewrite and refine these prompts to create more potent adversarial inputs, while defenders are simultaneously optimized to withstand these evolving threats and maintain task performance. To further stabilize training and foster coordination, we introduce a \textbf{public baseline} for advantage estimation: agents within the same functional group (e.g., attackers or defenders) share the group’s mean return as their baseline. This mechanism enables agents to learn from peer behaviors, reduces variance in policy updates, and strengthens intra-group cooperation. With training, attackers evolve beyond static attack templates, while defenders acquire more robust and generalizable safety behaviors. This co-evolutionary process drives continuous safety enhancement, mitigating the risk of overfitting to fixed attack distributions and enabling resilience against adaptive adversaries.

Experiments on three representative MAS attack scenarios—agent manipulation, message corruption, and user instruction hijacking—demonstrate the effectiveness of AdvEvo-MARL in enhancing system robustness. Further task benchmarks show minimal performance degradation, and in some cases even improved task capabilities, underscoring the potential of AdvEvo-MARL as a standardized framework for building MAS that are both safe and capable.

In summary, our main contributions are three-folds:
\begin{itemize}

    \item We propose AdvEvo-MARL, a novel multi-agent reinforcement learning framework that internalizes safety awareness within each agent through adversarial co-evolution. In this evolving paradigm, attackers and defenders iteratively compete and improve, leading to increasingly robust strategies on both sides.

    \item We introduce a public baseline mechanism for advantage estimation, where agents within the same functional group (e.g., attackers or defenders) use the group’s mean return as a baseline. This design promotes collaborative learning among agents and enables more stable policy updates during training.

    \item Experiments across multiple representative MAS attack settings demonstrate consistent safety gains—achieving up to a maximum of 18.33\% improvement. Further evaluations on standard task benchmarks reveal minimal degradation and, in several cases even enhanced task performance, underscoring AdvEvo-MARL’s effectiveness in simultaneously promoting multi-agent system safety and task utility.

\end{itemize}

\section{Related Work}
Our work builds on two main research lines. The first examines safety in MAS, where adversarial threats such as agent manipulation and message corruption motivates defenses like self-verification, guard agents, and peer inspection. The second explores multi-agent reinforcement learning (MARL), which has enabled coordinated training and has recently been applied to LLM-based systems. These perspectives motivate our proposed AdvEvo-MARL, which unifies safety and MARL by co-evolving attackers and defenders to embed intrinsic safety awareness into agents.

\subsection{Safety in Multi-Agent Systems.}
LLMs are known to exhibit safety vulnerabilities, especially when exposed to adversarial attacks. Equipping agents with external tools or memory systems further expands the attack surface \citep{raza2025trism, agentpoison}. While multi-agent systems (MAS) built upon such agents demonstrate strong task-solving capabilities, they are also vulnerable to a wide range of threats, most commonly: (1) manipulating agents to induce malicious behaviors \citep{ netsafe}, and (2) corrupting communication messages or workflow execution \citep{he2025red, breakingagents}. To mitigate these risks, several defense strategies have been proposed. Some works leverage \emph{self-verification}, encouraging each agent to assess the benignness of its inputs before responding \citep{peerguard, autoinject}, while others employ a dedicated \emph{guard agent} to monitor and rectify message flows \citep{ autoinject}. Another line of research collects safety-oriented interaction trajectories and trains graph neural networks to detect and correct unsafe responses \citep{gsafeguard}. Furthermore, decentralized defenses have also been explored, where agents inspect one another to form peer-based protection \citep{peerguard}.  
Although these approaches provide partial safeguards, they face key limitations. Individual agents often lack the capacity to detect sophisticated attacks, while centralized guard agents introduce a single point of failure and impose computational overhead in complex systems. In contrast, we advocate embedding safety awareness directly into each agent through reinforcement learning, enabling intrinsic defense capabilities and fundamentally improving the robustness of MAS.  
\subsection{Multi-Agent Reinforcement Learning.}
Reinforcement learning (RL) has proven effective in post-training LLMs \citep{shao2024deepseekmath, kimi, pan2025metaspatialreinforcing3dspatial}, with methods such as Proximal Policy Optimization (PPO) and Group Relative Policy Optimization (GRPO) yielding substantial performance gains \citep{shao2024deepseekmath}. More recently, RL has also been applied to enhance agentic behaviors in language-based systems \citep{jin2025search}.  
Multi-agent reinforcement learning (MARL), exemplified by algorithms like MAPPO and QMIX \citep{kang2023cooperative, rashid2020weighted}, extends RL to coordinated multi-agent settings \citep{liu2025llm}. Several recent studies adapt MARL to LLM-based systems: one line of work applies MARL to improve collaborative agent behaviors in structured game environments \citep{maporl}; another develops hierarchical MAS with high-level planners and low-level executors using parameter sharing to enhance meta-reasoning \citep{rema}; yet another treats each Retrieval-Augmented Generation (RAG) module as an agent, applying MARL to jointly optimize task performance \citep{chen2025improving}.  
\textbf{However}, most methods train a single backbone model with \textbf{shared parameters across agents} \citep{pan2025evomarlcoevolutionarymultiagentreinforcement}, limiting true agent-level diversity. In contrast, our framework trains \textbf{multiple distinct backbone models} collaboratively under RL, enabling genuine co-evolution. Building on these advances, our work explores MARL as a vehicle to improve MAS safety. By co-evolving attackers and defenders in an adversarial learning environment, we embed safety awareness directly into task agents through continuous interaction and adaptation, fostering robust and generalizable defense capabilities.

\section{Preliminary}

We formulate the interaction among learning agents as a partially observable Markov game:

\begin{equation}
    \begin{aligned}
        \mathcal{G} = (\mathcal{S}, \{\mathcal{A}_i\}_{i=1}^N, P, \{\mathcal{O}_i\}_{i=1}^N, \gamma,   \mathcal{T}),
    \end{aligned}
\end{equation}

where $\mathcal{S}$ denotes the state space, $\mathcal{A}_i$ represents the action space of agent $i$, $P$ is the state transition function, $\mathcal{O}_i$ is the observation function for agent $i$, $\mathcal{\gamma}$ is the discount factor, and $\mathcal{T}$ is the finite time horizon. Each agent $i \in \{1, \dots, N\}$ follows a stochastic policy $\pi_i(a_i \mid o_i)$, conditioned on local observation $o_i \sim \mathcal{O}_i(s)$, and jointly contributes to the environment evolution via the composite action $a = (a_1, \dots, a_N)$. 
In the context of LLMs-based agents, instead of treating each token as an action, we define the action of an agent as generating a complete response that consists of a token sequence.

The agents are partitioned into two disjoint sets: attackers $\mathcal{A}$ and defenders $\mathcal{D}$, $\mathcal{A} \cap \mathcal{D} = \emptyset$ and $\mathcal{A} \cup \mathcal{D} = \{1, \dots, N\}$.
The attackers attempt to compromise system's safety guardrail,
while the defenders must resist adversarial attacks and preserve task performance. All agents interact over the course of $T$ steps. At the end of each episode, the system produces a final output $y = \Phi(\tau)$, where $\tau = (s_0, a_0, s_1, \dots, s_T)$ denotes the complete trajectory induced by the multi-agent interaction. This output is then evaluated by the environment or a trusted judge to form a global reward $G(\tau)$, upon which each agent receives its own local reward $r_{i}$. The learning goal is to co-evolve attackers and defenders under shared dynamics and finally induce a stable and robust equilibrium between attacker and defender populations. This is captured by the following game-theoretic objective, where $\{{\pi_k}\}^N_{k = 1}$ denotes the joint policy of all N agents:
\begin{equation}
    \begin{aligned}
    \max _{\left\{\pi_j\right\}_{j \in \mathcal{D}}} \min _{\left\{\pi_i\right\}_{i \in \mathcal{A}}} \mathbb{E}_{\tau \sim\left\{{\pi_k}\right\}^N_{k = 1}}\left[\sum_{j \in \mathcal{D}} r_j(\tau)-\sum_{i \in \mathcal{A}} r_i(\tau)\right].        
    \end{aligned}
\end{equation}

\section{Methodology}

\begin{figure}[t] 
    \centering
    \includegraphics[draft=false, width=\textwidth]{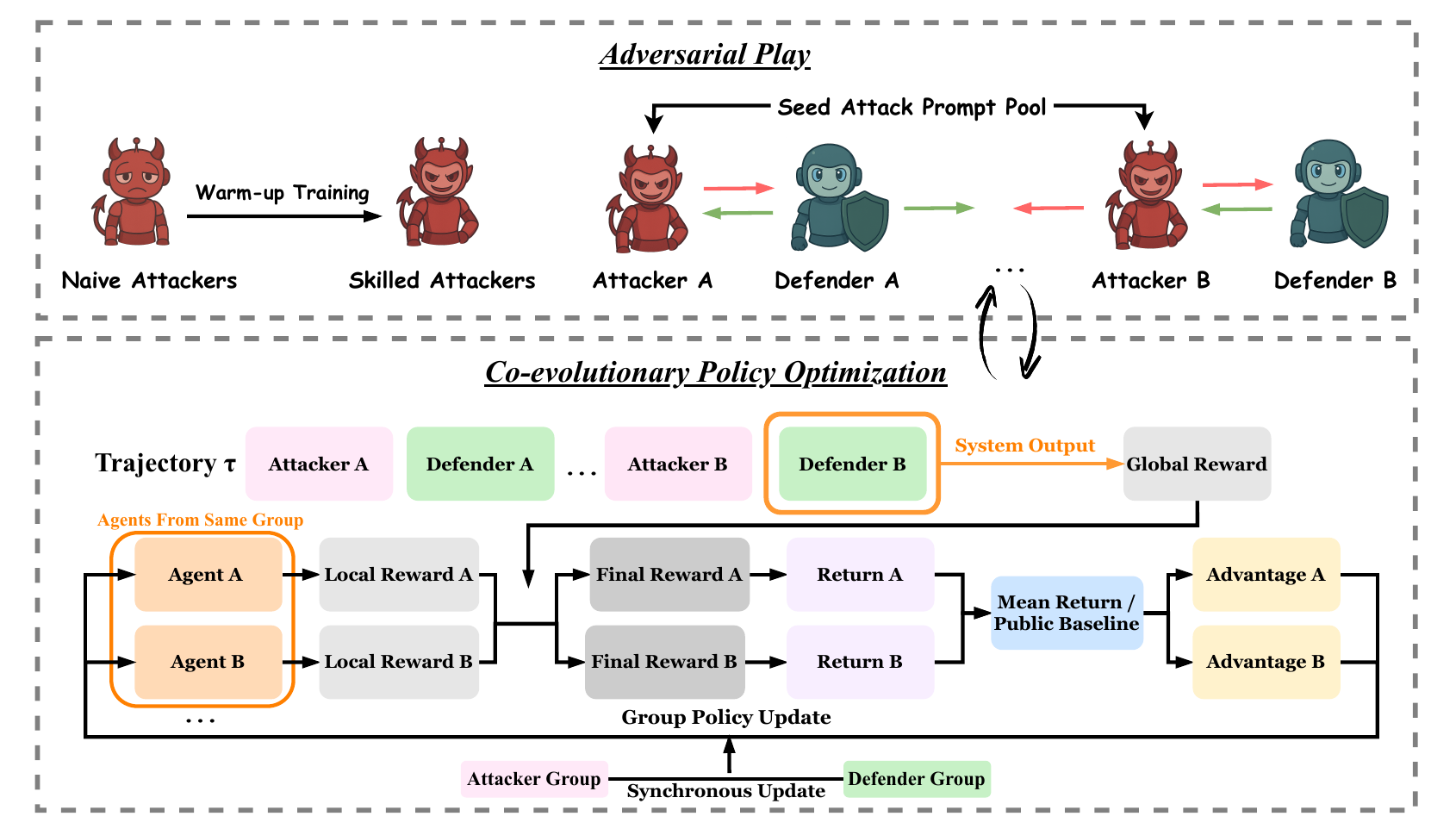}
    \caption{\textbf{Framework}. AdvEvo-MARL begins by warming up attacker agents through supervised fine-tuning to embed prior knowledge of jailbreak behaviors. Then, attackers and defenders learn to co-evolve via adversarial multi-agent reinforcement learning. During policy updates, agents within the same functional group (i.e., attackers or defenders) leverage a public baseline which is computed as the mean return of their respective group to estimate their individual advantages for optimization.}
    \label{fig:framework}
\end{figure}

In this section, we introduce AdvEvo-MARL, a multi-agent reinforcement learning framework designed to improve the safety of multi-agent systems. We first provide an overview of AdvEvo-MARL, then detail the attacker warm-up procedure, and finally present the adversarial RL pipeline with public-baseline-based advantage estimation.

\subsection{Overview}

As shown in \cref{fig:framework}, AdvEvo-MARL unfolds in two stages. First, an \emph{attacker warm-up} phase uses supervised fine-tuning to inject prior knowledge of jailbreak strategies, preventing trivial or ineffective attacks at the start of training. Upon this initialization, we introduce an \emph{adversarial co-evolutionary RL stage} where attackers and defenders are jointly optimized through repeated interactions, enabling defenders to acquire robust and adaptive safety behaviors against evolving threats. To stabilize learning and encourage group-consistent updates, agents within the same role leverage a \emph{public baseline} for advantage estimation, reducing variance and promoting effective collaboration.

\subsection{Bootstrapping Adversarial Generation via Attacker Warm-Up}

As attackers lack a prior understanding of jailbreak behaviors and adversarial prompting techniques, we first conduct warm-up training before MARL. 
We construct dataset $D_{adv}$ consisting of paired samples of the form ($x_{behavior}$, $x_{attack}$), where $x_{behavior}$ is the trivial harmful questions, and $x_{attack}$ is the re-written attack prompts using certain jailbreak techniques. Specifically, we begin by sampling 1,000 harmful behaviors from existing public datasets, ensuring broad coverage across diverse categories of harmful content. We then apply representative jailbreak strategies to generate corresponding adversarial attack prompts, obtaining an initial jailbreak prompt dataset $D_{init}$. Given the original questions and their associated attack variants, we employ an advanced reasoning model to synthesize multi-step reasoning traces that illustrate how to construct effective adversarial prompts.
To ensure quality, we filter out invalid reasoning trajectories that are contradictory, off-topic, or vague using a LLM-as judge method. The resulting dataset $D_{adv}$ contains approximately 4,000 high-quality training samples. AdvEvo-MARL leverages imitation learning to equip attackers with jailbreak knowledge from the curated $D_{adv}$, thereby accelerating exploration in the early stages of training.

\subsection{AdvEvo-MARL: Safe and Capable Multi-Agent Systems via co-evo RL}

To build a safe and capable MAS, we embeds safety awareness directly into agents through adversarial co-evolution, enabling them to withstand evolving attacks while maintaining strong task performance. Importantly, we \textbf{trains multiple backbone models} collaboratively under RL, \textbf{rather than} relying on a \textbf{single shared-parameter model}, ensuring genuine co-evolution across diverse agents.

\paragraph{Training Algorithm}

Following the attacker warm-up stage, both attackers and defenders are jointly optimized within a co-evolutionary multi-agent reinforcement learning process. All agents are trained using REINFORCE++ to improve both system safety and task performance \citep{hu2025reinforce++}. To facilitate collaborative learning and stabilize policy updates, we introduce a public baseline for advantage estimation. 

Specifically, during each rollout episode, the advantage for each agent is computed relative to the mean return of all agents within the same role group (i.e., attackers or defenders), rather than being estimated solely from its own return trajectory.
Formally, for episode $\tau$ we define:

\begin{equation}
    \begin{aligned}
        b^A(\tau) = \frac{1}{|\mathcal A|} \sum_{i\in\mathcal A} r^A_i(\tau), 
\qquad
b^D(\tau) = \frac{1}{|\mathcal D|} \sum_{j\in\mathcal D} r^D_j(\tau),
    \end{aligned}
\end{equation}

where $b^A$ and $b^D$ denote the mean return value of attackers and defenders respectively.
The resulting advantage estimate for any agent $k \in \{1, \dots, N\}$ is then given by

\begin{equation}
    \begin{aligned}
    \hat A_k(\tau) = r_k(\tau) -
\begin{cases}
b^A(\tau), & \text{if } k\in\mathcal A, \\
b^D(\tau), & \text{if } k\in\mathcal D .
\end{cases}        
    \end{aligned}
\end{equation}

Finally, the training loss for agent $k$ is defined as:

\begin{equation}
    \begin{aligned}
    \mathcal{L}_{\mathrm{REINFORCE++}}\left(\theta_k\right)= & -\mathbb{E}_t\left[\min \left(r_{t, k}\left(\theta_k\right) \hat{A}_{t, k}, \operatorname{clip}\left(r_{t, k}\left(\theta_k\right), 1-\varepsilon, 1+\varepsilon\right) \hat{A}_{t, k}\right)\right] \\
& +\beta_{\mathrm{KL}} \mathbb{E}_t\left[\operatorname{KL}\left(\pi_{\theta_k}\left(\cdot \mid x_{t, k}\right) \| \pi_{\mathrm{ref}, k}\left(\cdot \mid x_{t, k}\right)\right)\right],       
    \end{aligned}
\end{equation}

where $r_{t,k}(\theta_k) = \tfrac{\pi_{\theta_k}(a_{t,k}\mid x_{t,k})}{\pi_{\theta_k^{\text{old}}}(a_{t,k}\mid x_{t,k})}$ denotes the importance sampling ratio, clipping $clip$ restricts updates magnitude, and the KL term measures divergency between learned policy $\pi_{\theta_k}$ and reference policy $\pi_{ref, k}$ to regulate training.

\paragraph{Reward Modeling}

To care distinct objectives of attackers and defenders, we design separate reward mechanisms for each agent type. Attackers receive rewards based on whether the final system output achieves the intended malicious goal, as evaluated by a global reward signal. In contrast, defenders are responsible for both resisting jailbreak attempts and fulfilling their assigned tasks. Relying solely on the global reward, however, can introduce misaligned incentives for defenders: an individual agent may receive misleading feedback due to the behavior of others. For instance, when some agents generate unsafe responses but the aggregated system output remains benign.

To address this issue, we assign rewards based on both individual agent's response and the final system output. Therefore, the rewards of defenders are evaluated at both the local response level and the global system level, as a combination of task performance and safety compliance.
All agents also receive a formatting reward that enforces their outputs to put reasoning process between \texttt{<think>} and \texttt{</think>} and enclose final response with \texttt{<response>} and \texttt{</response>} tags.
The overall reward is formulated as:

\begin{equation}
    \begin{aligned}
    R_k =
\begin{cases}
\displaystyle \gamma_f \cdot f \;-\; \alpha_s \cdot s, & \text{if } k \in \mathcal{A}, \\[1em]
\displaystyle \alpha_s \cdot s \;+\; \beta_t \cdot t \;+\; \gamma_f \cdot f, & \text{if } k \in \mathcal{D},
\end{cases}
    \end{aligned}
\end{equation}

where $s$, $t$, and $f$ represent the rewards for safety, task utility, and format compliance respectively. For both safety and task performance, a reward of $1$ is assigned if the output is safe or correct, and $-1$ otherwise. For formatting, a reward of $0.5$ is given if the response satisfies the pre-defined structure, and $-0.1$ otherwise.
In practice, we prioritize safety in the first half of training ($\alpha_s = 1$, $\beta_t = 0.5$), and reverse the weights afterward to emphasize task performance.

\section{Experiments}
Experiments cross 3 representative multi-agent attack scenarios and 3 task-specific benchmarks to assess its effectiveness in enhancing both safety and task utility. We first describe the experimental setup. Then we report results on red team attacks to demonstrate the robustness of our approach against adversarial threats. Next, we present task evaluations to assess the model's general task performance. Finally, we conduct ablation studies to validate the design choices of AdvEvo-MARL.

\subsection{Experimental Setup}

\textbf{Multi-agent systems.}
To ensure a comprehensive evaluation under varying communication structures, we consider three representative system topologies in our experiments. (1) Chain mode: agents interact sequentially. Each agent can only observe the message from its immediate predecessor. (2) Tree mode: a hierarchical structure where two child agents exchange messages and a parent agent summarizes the communication history to produce a final output. (3) Complete mode: a fully connected topology where each agent can send and receive messages to and from all other agents. All experiments are conducted with three agents. Unless otherwise specified, we use QWen2.5 instruction-tuned models (3B and 7B) as the backbone foundation models.

\textbf{Attack methods.}
We choose three widely adopted attack strategies mainly focusing on jailbreak attacks and harmful information propagation within MAS. (1) \emph{NetSafe} \citep{netsafe}, alters agent behavior by injecting ‘dark traits’ into profile configurations. One agent is randomly selected as malicious attacker in each episode. (2) \emph{AutoInject} \citep{autoinject}, randomly injects adversarial prompts into communication messages between agents. (3) UserHijack, manipulate user instructions to insert targeted adversarial content, simulating compromised user input. 

\textbf{Baselines.}
We compare AdvEvo-MARL against several baseline methods. (1) Vanilla instruction-tuned QWen2.5 series 3B and 7B models without safety training as backbone models. (2) Challenger,     a self-verification strategy where individual agent verify the benignness of its input before generating a response. All agents are equipped with this self-checking mechanism in our setting. (3) Inspector, introduces external guard agents to detect and correct malicious messages. We only deploy one inspector agent to monitor all message flows during interaction.

\textbf{Datasets.} 
In training, we sample 4,000 problems from levels 3–5 of MATH-500 dataset \citep{math} to serve as regular task prompts for defenders. In adversarial training, we use the described seed pool for attack rewriting.
For \textbf{system safety evaluation}, we follow the original NetSafe protocol and adopt its official evaluation dataset. Meanwhile, we construct a 300 adversarial prompts pool by sampling JailbreakBench, Wild Jailbreak, and Strong Reject. These prompts are used in both AutoInject and UserHijack settings.
For \textbf{general task evaluation}, we select 3 prevailing benchmarks: (1) \emph{mathematical reasoning}: AIME'24 \& AIME'25 \citep{aime}, challenging high-school mathematics requiring deep thinking and creative problem-solving, each containing 30 questions in total; (2) \emph{coding}: LiveCodeBench (v6, 2025.01 - 2025.05) \citep{livecodebench}, collecting coding problems from live online platforms, providing a realistic, dynamic, challenging environment for coding capability evaluation; (3) \emph{general reasoning}: GPQA-diamond \citep{gpqa-diamond}, 100 graduate-level Q\&A problems encompassing physics, chemistry, biology and other scientific domains.  

\textbf{Metrics.} We employ three metrics to comprehensively evaluate both the robustness and utility of multi-agent systems. (1) Attack success rate (ASR): the proportion of  evaluation samples where the system ultimately produces a harmful response. (2) Contagion rate (PR): the ratio of agents that exhibit unsafe behaviors at any point during the interaction episode, reflecting the system's process-level safety. (3) Task performance: we adopt accuracy (Acc) for mathematical and general reasoning tasks, and Pass@1 for coding tasks.

\subsection{Main Results}

\begin{table}[t]
\centering
\caption{Attack success rate (ASR) and contagion rate (CR) on NetSafe, AutoInject, and UserHijack attack scenarios across chain, tree, and complete graph topology systems. Lower ASR and CR indicate stronger robustness. Best-performing result is highlighted in \textbf{bold} and second-best is \underline{underlined}.}
\vspace{6pt}
\label{tab:main_results_asr_cr}
\resizebox{\linewidth}{!}{%
\begin{tabular}{ll*{2}{c}*{6}{c}*{6}{c}}
\toprule
\multicolumn{2}{c}{\multirow{2}{*}{}} &
\multicolumn{2}{c}{\multirow{2}{*}{\textbf{NetSafe}}} &
\multicolumn{6}{c}{\textbf{AutoInject}} &
\multicolumn{6}{c}{\textbf{UserHijack}} \\
\cmidrule(lr){5-10}\cmidrule(lr){11-16}
&&&&
\multicolumn{2}{c}{\textbf{AIME}} &
\multicolumn{2}{c}{\textbf{GPQA}} &
\multicolumn{2}{c}{\textbf{LiveCodeBench}} &
\multicolumn{2}{c}{\textbf{AIME}} &
\multicolumn{2}{c}{\textbf{GPQA}} &
\multicolumn{2}{c}{\textbf{LiveCodeBench}} \\
\cmidrule(lr){5-6}\cmidrule(lr){7-8}\cmidrule(lr){9-10}
\cmidrule(lr){11-12}\cmidrule(lr){13-14}\cmidrule(lr){15-16}
&& \textbf{ASR} & \textbf{CR} &
\textbf{ASR} & \textbf{CR} &
\textbf{ASR} & \textbf{CR} &
\textbf{ASR} & \textbf{CR} &
\textbf{ASR} & \textbf{CR} &
\textbf{ASR} & \textbf{CR} &
\textbf{ASR} & \textbf{CR} \\
\midrule
\multirow{10}{*}{Chain}
& GPT-3.5 & 10.89\% & 11.88\% & 3.33\% & 3.89\% & 3.03\% & 3.7\% & 5.14\% & 5.62\% & 15\% & 15.56\%    & 10.61\% & \underline{6.79\%}    & 19.24\% & 17.76\% \\
& GPT-4o-mini & \textbf{0\%} & \textbf{0\%} & 3.33\% & 3.33\% & 5.05\% & 5.39\% & \underline{1.14\%} & \underline{1.14\%}     & \underline{3.33\%} & \underline{7.78\%}   & \underline{4.55\%} & 8.67\%    & \textbf{2.29\%} & \textbf{5.64\%} \\
\cmidrule(lr){2-16}
& Vanilla-3B & 11.88\% & 36.14\% & 15\% & 19.44\% & 19.7\% & 22.05\% & 16\% & 16.57\%    & 33.33\% & 37.78\%    & 24.24\% & 32.59\%    & 26.29\% & 35.27\% \\
& Vanilla-7B- & 21.78\% & 40.35\% & 13.33\% & 15\% & 21.21\% & 21.63\% & 7.43\% & 8.76\%   & 25.58\% & 25.58\%    & 17.68\% & 21.64\%    & 22.29\% & 28.53\% \\
& Challenger-3b & 8.91\% & 17.57\% & 13.33\% & 16.39\% & 20.2\% & 20.54\% & 12.57\% & 15.81\%   & 16.67\% & 19.43\%    & 25.25\% & 28.74\% & 24\% & 21.65\% \\
& Inspector-3b & 1.98\% & \underline{2.23\%}     & \underline{1.67\%} & 1.67\% & 3.03\%       & 3.28\% & 4.57\% & 4.57\% & 15\% & 9.44\%  & 11.11\% & 18.25\%  & 14.29\% & 16.44\% \\
& Challenger-7b & 3.96\% & 9.24\% & 8.33\% & 8.33\% & 13.64\% & 14.93\% & 19.43\% & 21.14\%   & 16.67\% & 17.69\%  & 16.16\% & 12.35\% & 14.29\% & 18.24\% \\
& Inspector-7b & 1.98\% & 2.72\%   & \textbf{0\%} & \textbf{0\%}        & 3.54\% & 4.38\% & 2.29\% & 2.57\% & 13.33\% & 10.66\%    & \underline{4.55\%} & 6.93\%   & 8\% & 7.59\% \\
\cmidrule(lr){2-16}
& AdvEvo-MARL-3B (ours) & 6.93\% & 35.64\%   & \textbf{0\%} & \underline{1.11\%}              & \underline{1.52\%} & \underline{2.19\%}      & 2.29\% & 2.29\% & 8.33\% & 9.44\% & 7.07\% & 8.25\% & 8.29\% & 15.05\% \\
& AdvEvo-MARL-7B (ours)  & \underline{0.99\%} & 19.14\%    & \textbf{0\%} & \underline{1.11\%}      & \textbf{0.51\%} & \textbf{1.85\%}     & \textbf{0.57\%} & \textbf{0.19\%}     & \textbf{1.67\%} & \textbf{0.56\%}      & \textbf{4.04\%} & \textbf{3.03\%}       & \underline{6.86\%} & \underline{6.29\%} \\
\midrule
\multirow{10}{*}{Tree}
& GPT-3.5 & 8.91\% & 9.9\%    & \textbf{0\%} & \textbf{0\%}     & \textbf{0\%} & \textbf{0.25\%}    & \underline{1.71\%} & 1.71\%      & 10\% & 17.5\% & 9.6\% & 16.67\% & 15.43\% & 24.57\% \\
& GPT-4o-mini & \textbf{0\%} & \textbf{0\%}   & 10\% & 2.5\%   & \textbf{0\%} & \underline{0.38\%}    & 3.43\% & \underline{1.57\%}     & 6.67\% & \textbf{2.92\%}      & \underline{4.55\%} & \textbf{2.15\%}       & \underline{6.29\%} & \textbf{3.71\%} \\
\cmidrule(lr){2-16}
& Vanilla-3B & 27.72\% & 41.34\% & 18.33\% & 13.75\% & 13.64\% & 9.34\% & 11.43\% & 10.29\% & 35\% & 41.25\% & 33.84\% & 42.68\% & 29.71\% & 37.43\% \\
& Vanilla-7B & 16.83\% & 26.98\% & 31.67\% & 22.5\% & 37.37\% & 22.6\% & 22.86\% & 18.29\% & 35\% & 33.33\% & 26.26\% & 28.03\% & 29.14\% & 29\% \\
& Challenger-3b & 22.77\% & 38.61\% & 3.33\% & 1.67\% & 4.04\% & 1.77\% & 4.57\% & 3\% & 30\% & 36.67\% & 25.76\% & 40.4\% & 26.29\% & 37.86\% \\
& Inspector-3b & 8.91\% & \underline{13.86\%}    & 10\% & 9.58\% & 3.54\% & 4.04\% & 4.57\% & 5.43\% & 10\% & 30.83\% & 9.6\% & 30.3\% & 10.29\% & 28.86\% \\
& Challenger-7b & 38.61\% & 51.49\% & 13.33\% & 7.92\% & 13.64\% & 9.09\% & 17.14\% & 10.29\% & 30\% & 34.58\% & 24.75\% & 26.89\% & 21.71\% & 27\% \\
& Inspector-7b & 8.91\% & 12.87\%    & 3.33\% & 4.58\%     & 4.04\% & 3.79\%   & \underline{1.71\%} & 3.86\%     & \underline{5\%} & 21.25\%     & 10.61\% & 20.45\%    & 10.29\% & 22\% \\
\cmidrule(lr){2-16}
& AdvEvo-MARL-3B (ours) & \underline{1.98\%} & 34.98\%    & \textbf{0\%} & \underline{0.56\%}    & \underline{1.01\%} & 2.86\%    & \underline{1.71\%} & 2.29\%    & 8.33\% & 16.11\% & \textbf{4.04\%} & 11.45\% & 9.25\% & 23.24\% \\
& AdvEvo-MARL-7B (ours) & 6.89\% & 24.42\%   & \underline{1.67\%} & 3.89\%     & \underline{1.01\%} & 2.02\%    & \textbf{0\%} & \textbf{0.19\%}    & \textbf{0\%} & \underline{4.44\%}     & 5.05\% & \underline{7.24\%}    & \textbf{5.14\%} & \underline{9.9\%} \\
\midrule
\multirow{10}{*}{Complete}
& GPT-3.5 & \textbf{0\%} & 21.53\%    & \textbf{0\%} & \underline{22.5\%}       & \textbf{0\%} & \textbf{0.38\%}       & \underline{1.14\%} & \underline{2.57\%}         & 26.67\% & 42.92\% & 18.18\% & 37.5\% & 34.86\% & 51.29\% \\

& GPT-4o-mini & \textbf{0\%} & \textbf{1.24\%}  & \textbf{0\%} & \textbf{0.42\%}      & \underline{0.51\%} & \underline{0.51\%}       & \textbf{0\%} & \textbf{0.29\%}     & \textbf{0\%} & \textbf{0.83\%}      & \textbf{0.51\%} & \textbf{1.89\%}    & \textbf{1.14\%} & \textbf{2.43\%} \\
\cmidrule(lr){2-16}
& Vanilla-3B & 42.57\% & 54.7\%    & 36.67\% & 71.11\%    & 16.06\% & 37.27\% & 26.86\% & 42.57\%        & 26.67\% & 53.33\% & 30.81\% & 56.82\% & 36\% & 65.43\% \\
& Vanilla-7B & 33.66\% & 43.07\%   & 33.33\% & 67.78\%   & 31.31\% & 43.68\% & 25.14\% & 39.29\%    & 28.33\% & 48.75\% & 29.29\% & 48.99\% & 34.86\% & 60.86\% \\

& Challenger-3B & \textbf{0\%} & 24.5\%        & 30.69\% & 61.67\% & 7.58\% & 29.55\% & \underline{1.14\%} & 14.43\%     & 40\% & 58.33\% & 34.85\% & 65.53\% & 33.14\% & 57.29\% \\
& Inspector-3B & 3.96\% & 27.23\%     & \underline{3.33\%} & 30.83\%     & 6.03\% & 26.89\%    & 4.57\% & 18.14\%   & 8.33\% & 48.75\%    & \underline{10.61\%} & 50.25\% & 19.14\% & 53.14\% \\
& Challenger-7B    & \textbf{0\%} & 22.28\%     & 15\% & 58.89\% & 22.73\% & 37.5\% & 15.43\% & 18.86\% & 38.33\% & 53.33\% & 33.84\% & 50.38\% & 29.71\% & 46\% \\
& Inspector-7B & 2.97\% & 17.82\%    & 5\% & 47.78\%     & 5.57\% & 29.47\%    & 4\% & 17.29\%     & 8.33\% & 41.25\%    & 16.57\% & 37.63\%     & 17.86\% & 39.57\% \\
\cmidrule(lr){2-16}
& AdvEvo-MARL-3B (ours) & \underline{0.27\%} & 3.65\%      & 6.73\% & 50.42\%     & 5.05\% & 26.26\%    & 3.43\% & 13.14\%      & \underline{4\%} & 47.22\%         & 17.68\% & 33.7\%    & 16.29\% & \underline{34.86\%} \\
& AdvEvo-MARL-7B (ours) & \textbf{0\%} & \underline{2.58\%}       & \underline{3.33\%} & 45.22\%     & 7.07\% & 29.46\%    & \underline{1.14\%} & 6.57\%      & 6.67\% & \underline{36.11\%}     & 11.11\% & \underline{28.48\%}    & \underline{14.86\%} & 40.48\% \\
\bottomrule
\end{tabular}}
\end{table}

Table ~\ref{tab:main_results_asr_cr} presents a comprehensive comparison of ASR and CR across a range of models, system topologies, and adversarial settings.
Among all open-source baselines, AdvEvo-MARL consistently achieve the lowest ASR and CR across nearly all configurations, demonstrating superior robustness against adversarial compromise in multi-agent systems.

Specifically, our models maintain ASR consistently below 10\% in simpler topology systems such as chain and tree, and remain competitive even in the more challenging complete graph topology, with a maximum ASR of 17.68\%, where high interconnectivity greatly facilitates adversarial propagation. In contrast, other open-source baselines frequently fail to maintain low ASR across all topologies. For example, in the tree setting, some models experience up to 38.61\% system-level compromise, and in the complete graph setup, ASR can rise as high as 65.53\%. In certain cases, these models even underperform relative to their vanilla counterparts: under the UserHijack setting, Challenger-7B reaches 38.33\% ASR, a 10\% increase over its non-defended variant. 
Notably, in chain and tree topologies, our models achieve low or even near-zero ASR, often matching or outperforming proprietary models (e.g., GPT-4o-mini).

Another key observation is that AdvEvo-MARL maintains low ASR while significantly suppressing CR even as the adversarial setting becomes more aggressive and the communication topology more interconnected. In contrast, many open-source defended baselines exhibit moderate ASR but much higher CR — often ranging from 30\% to even 60\% in densely connected environments, suggesting insufficient coordination or internal consistency when faced with adversarial attack contagion. Yet our models strive to retain CR below 35\% across all evaluation settings. This highlights AdvEvo-MARL’s superiority not only to improve individual agents' safety awareness, but also to facilitate collaboration among agents to disrupt adversarial attack spread across the system.

\begin{wrapfigure}[15]{l}{0.43\textwidth}  
  \centering
  \includegraphics[width=0.41\textwidth]{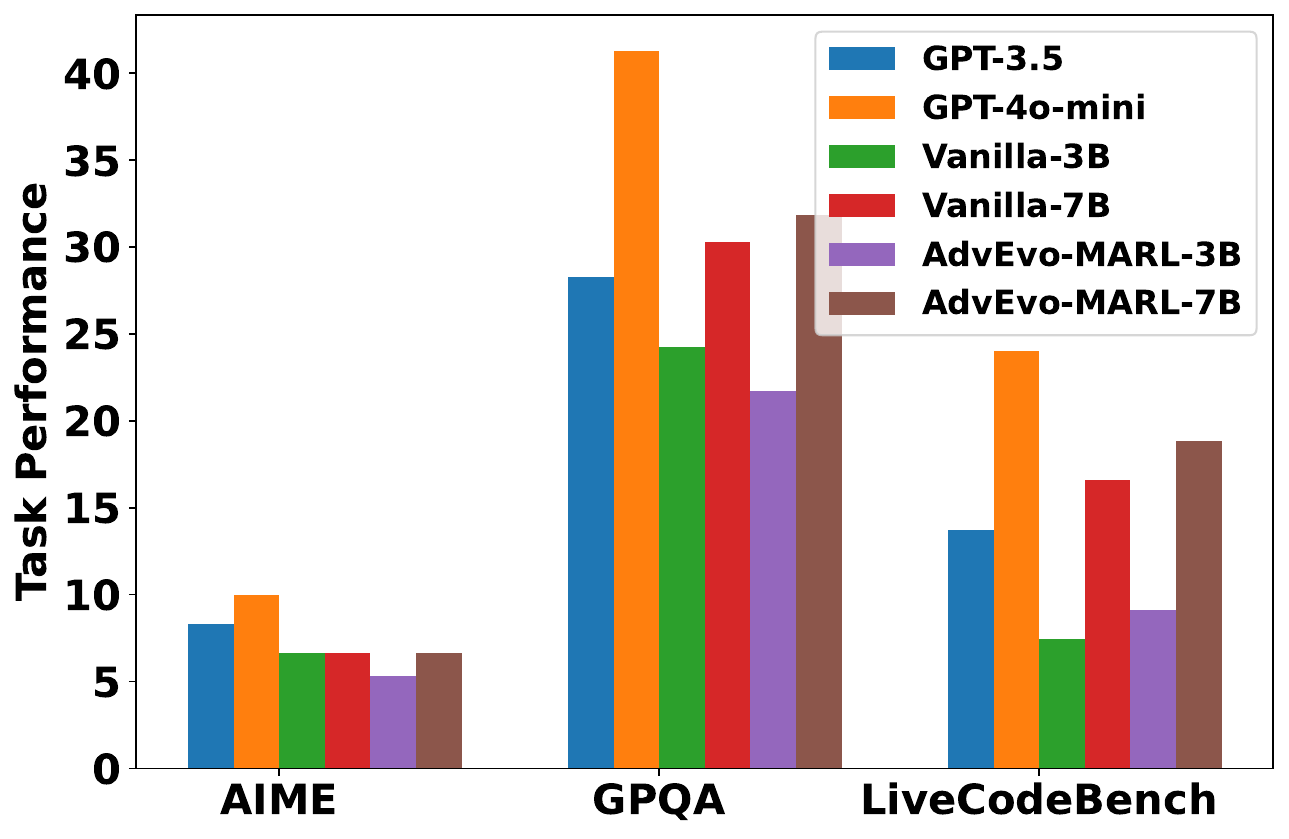}
  \caption{Task benchmark performance. AdvEvo-MARL exhibits minimal degradation and even improved results.}
  \label{fig:main results acc}
\end{wrapfigure}

We further evaluate the impact of AdvEvo-MARL on the system’s task capabilities across three representative benchmarks. Experimental results in \cref{fig:main results acc} show that our models retain strong task performance, with only a maximum 3\% accuracy drop observed among the 3B variants. Notably, the AdvEvo-MARL-7B model being trained exclusively on mathematical tasks, not only preserves its original task competence but even \emph{outperforms} its vanilla counterpart across all datasets, especially those deemed as out-of-distribution.
These findings provide clear evidence that safety-oriented training can be achieved without definitely sacrificing task ability. AdvEvo-MARL enables the development of agents that are both robust and performant, underscoring its potential as a principled framework for building safe yet capable multi-agent systems.

\subsection{Dynamic attacks and collaborative defense}
\vspace{-0.05in}

\begin{figure}[t]
  \centering
  \includegraphics[draft=false, width=\textwidth]{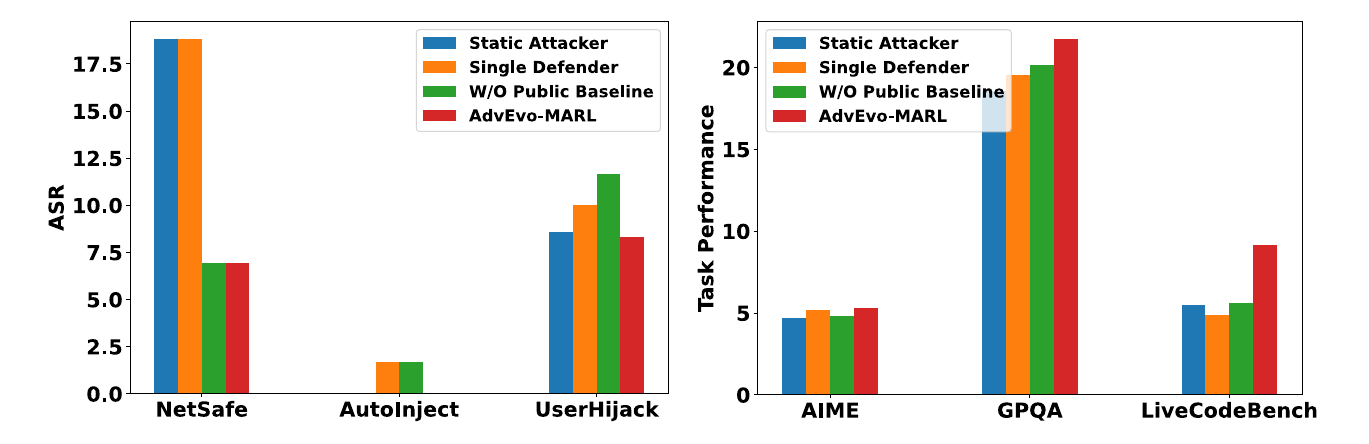}
  \caption{Performance variations under different training configurations. \emph{Left:} robustness performance, AdvEvo-MARL consistently maintains the lowest ASR, \emph{Right:} task performance, AdvEvo-MARL improves task utility across all settings, reaching a maximum 4\% gain on LiveCodeBench.}
  \label{fig:ablation asr acc}
\end{figure}

To evaluate the effectiveness of dynamic attacker modeling, we compare our MARL-based attacker framework with a static attacker baseline. In the static setting, adversarial prompts are drawn from a fixed pool without adaptation. In contrast, our method enables attackers to continuously generate and refine attack prompts through co-evolution with defenders.
As shown in \cref{fig:ablation asr acc}, our MARL-based attacker achieves significantly lower ASR under NetSafe threat, revealing a 12\% reduction, indicating that defenders trained with evolving attackers exhibit superior robustness. In the AutoInject and UserHijack settings, AdvEvo-MARL also yields marginally lower ASR, suggesting consistent safety improvements across threat models. 
Evaluations on task datasets also reveal that AdvEvo-MARL outperforms the static attacker baseline across all settings, achieving a maximum 4\% performance gain. These results suggest that the presence of dynamic attackers can encourage defenders to develop generalizable task-solving capabilities, highlighting the dual benefits of AdvEvo-MARL for enhancing both safety and utility.

\begin{wrapfigure}[14]{l}{0.48\textwidth} 
  \centering
  \includegraphics[width=0.46\textwidth]{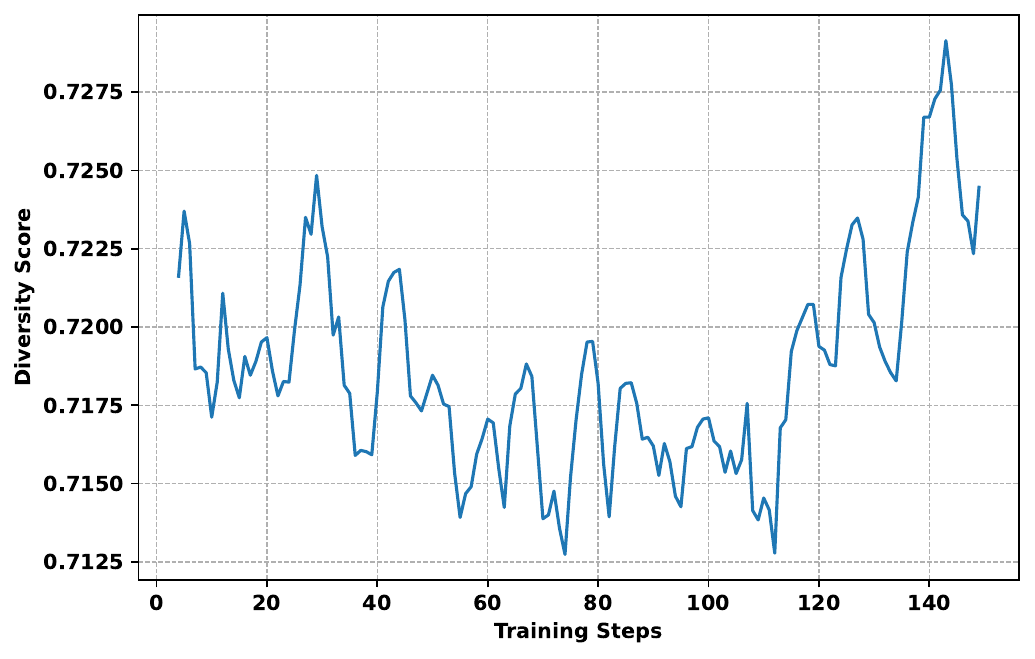}
  \caption{Attacker-generated prompts Diversity.}
  \label{fig: attacks diversity}
\end{wrapfigure}

We further investigate how our dynamic attacker evolves throughout the MARL training process. 
To quantify this progression, we measure the semantic similarity between generated attack prompts and all seed attacks to obtain diversity scores.
Notably, as shown in Figure~\ref{fig: attacks diversity}, the diversity of adversarial prompts generated by the attacker, reveals a non-monotonic but ultimately increasing trend over the course of training.
Despite fluctuations in early stages, the diversity steadily increases in the later phase, indicating that the attackers learns to produce increasingly varied and novel jailbreak attacks. 
This increased diversity coincides with enhanced robustness in the trained defenders which suggests a causal link. 
As attackers evolve and diversify, defenders are less likely to overfit and more capable of generalizing to previously unseen threats.
These results underscore that training with a dynamic attacker not only produces stronger adversarial prompts but also drives the emergence of more resilient and generalizable defense behaviors.

Another interesting question is whether training defenders in a MAS setting yields benefits over training them individually. Following the setup above, the empirical results in \cref{fig:ablation asr acc} demonstrate that AdvEvo-MARL exhibits the highest system safety and task utility across all evaluated settings. Notably under the NetSafe scenario, our models achieve a 12\% gain in robustness and a prominent 4\% enhancement in task utility comparatively. These improvements can be attributed to the emergence of collaborative defense behaviors that arise only through joint training. Such coordination and mutual adaptation among agents are difficult to achieve when agents are trained in isolation.

\subsection{Public Baseline based multi-agent reinforcement learning}

\begin{figure}[htbp]
    \centering
    
    \begin{subfigure}{0.32\textwidth}
        \includegraphics[width=\linewidth]{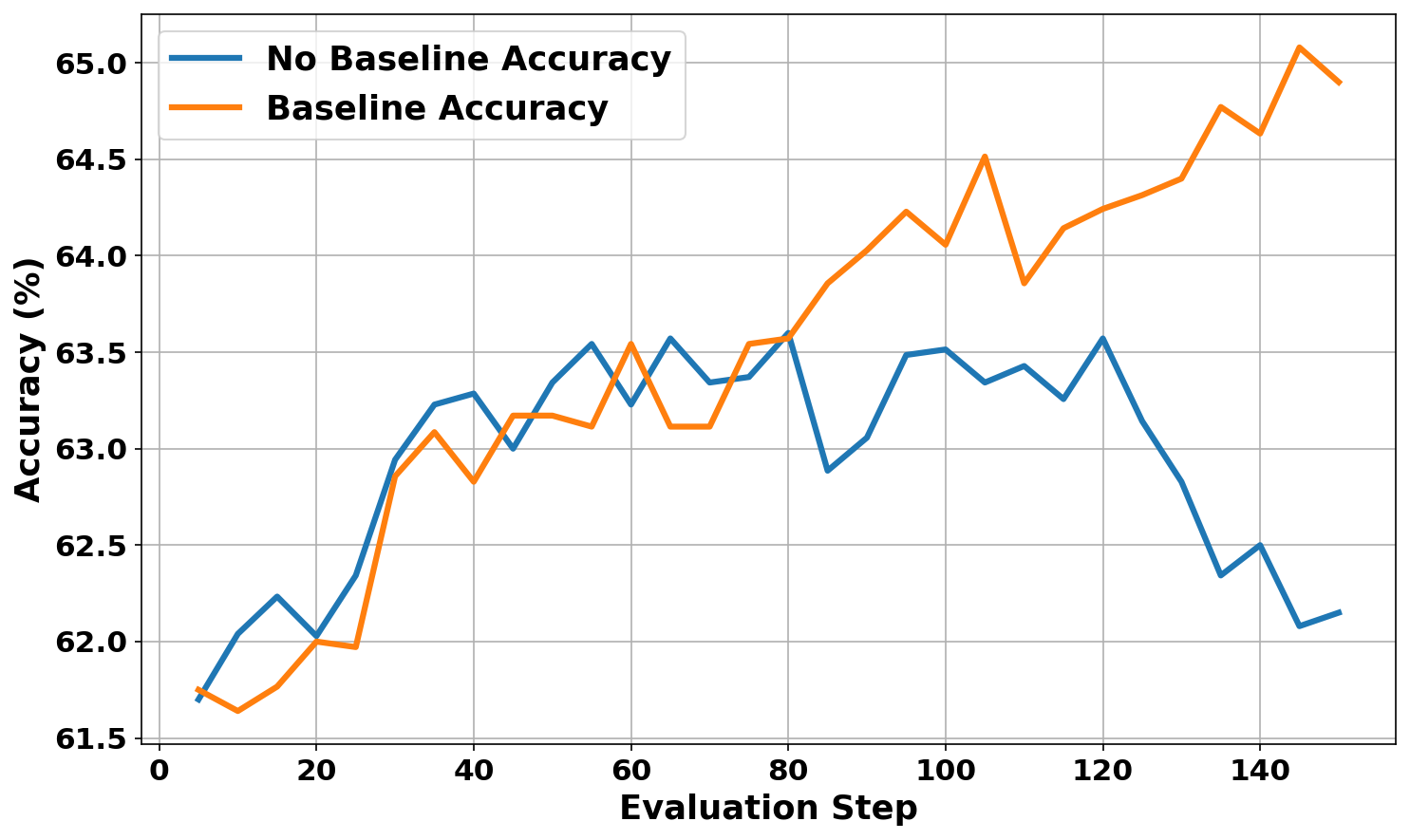}
        \caption{Evaluation accuracy.}
    \end{subfigure}
    \hfill
    \begin{subfigure}{0.32\textwidth}
        \includegraphics[width=\linewidth]{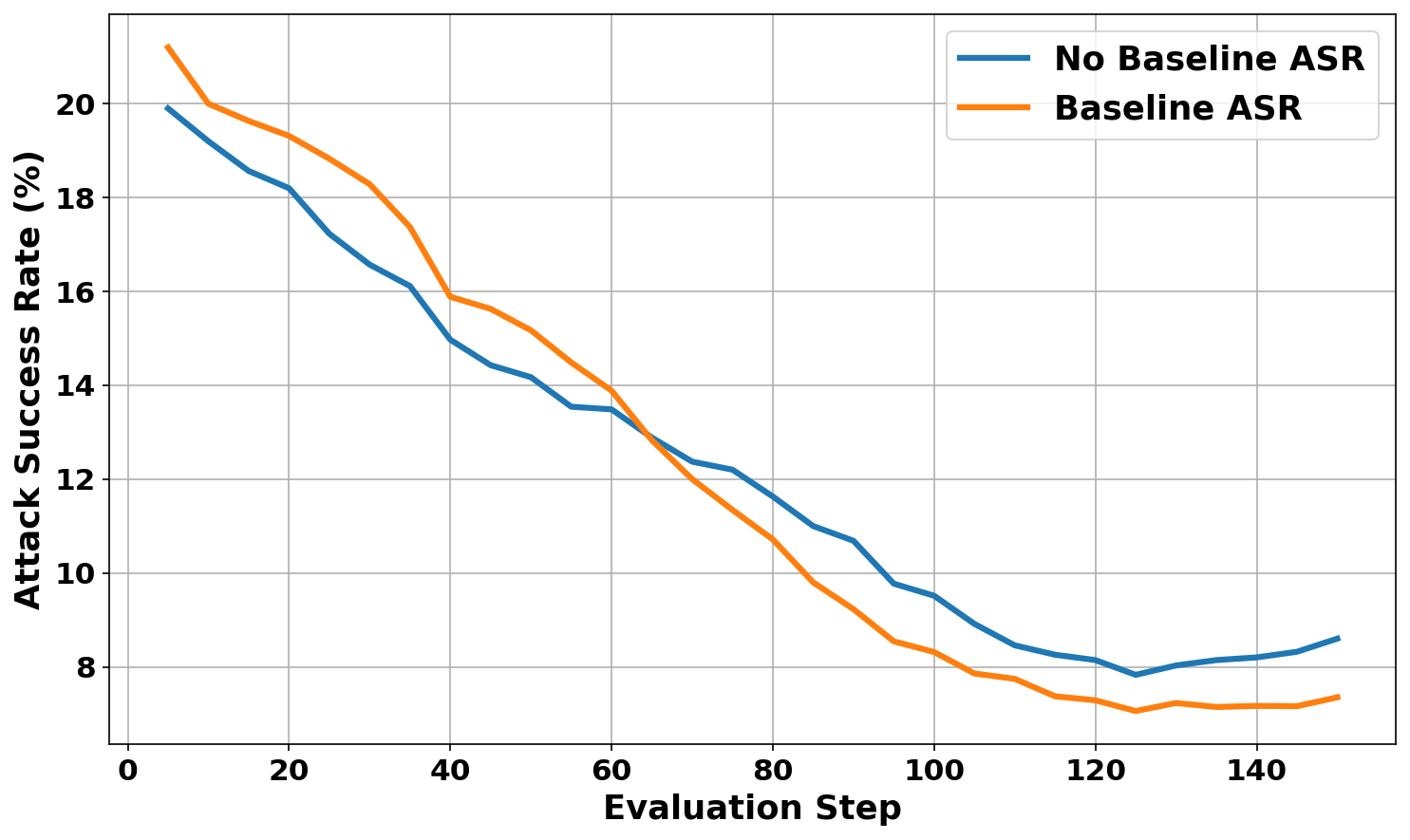}
        \caption{Attack success rate.}
    \end{subfigure}
    \hfill
    \begin{subfigure}{0.32\textwidth}
        \includegraphics[width=\linewidth]{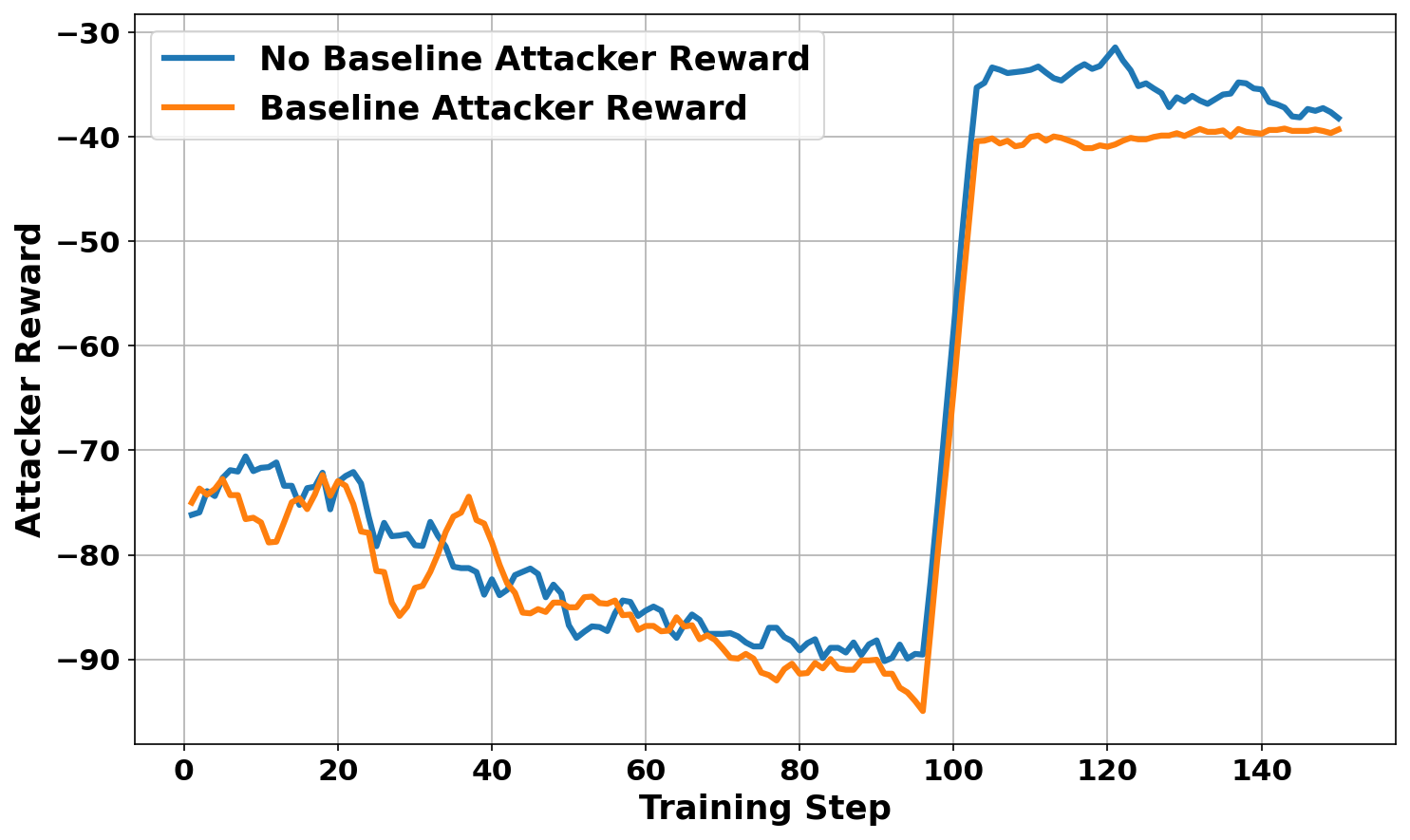}
        \caption{Attacker Reward.}
    \end{subfigure}
    
    \vskip\baselineskip
    
    \begin{subfigure}{0.45\textwidth}
        \includegraphics[width=\linewidth]{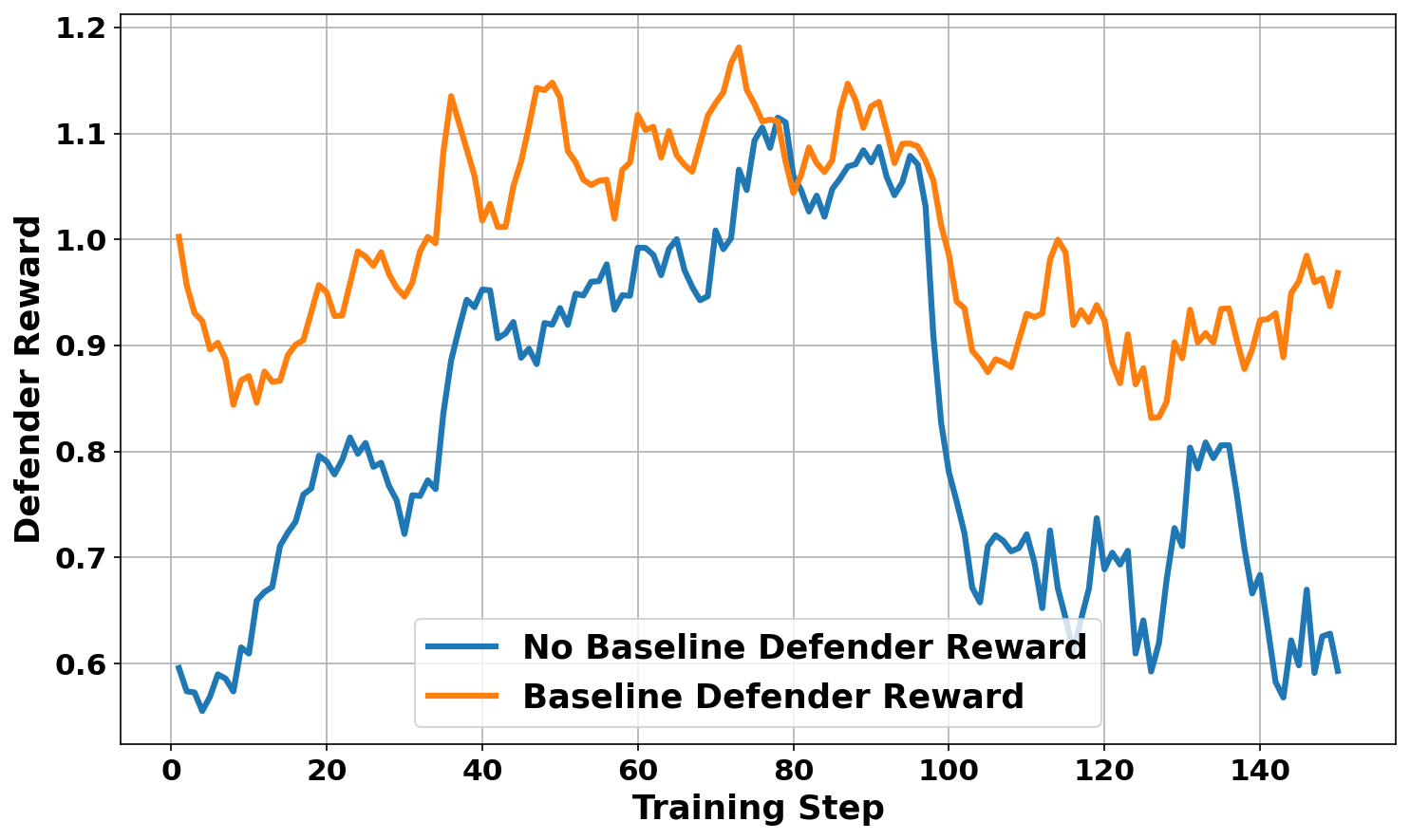}
        \caption{Defender Reward.}
    \end{subfigure}
    \hfill
    \begin{subfigure}{0.45\textwidth}
        \includegraphics[width=\linewidth]{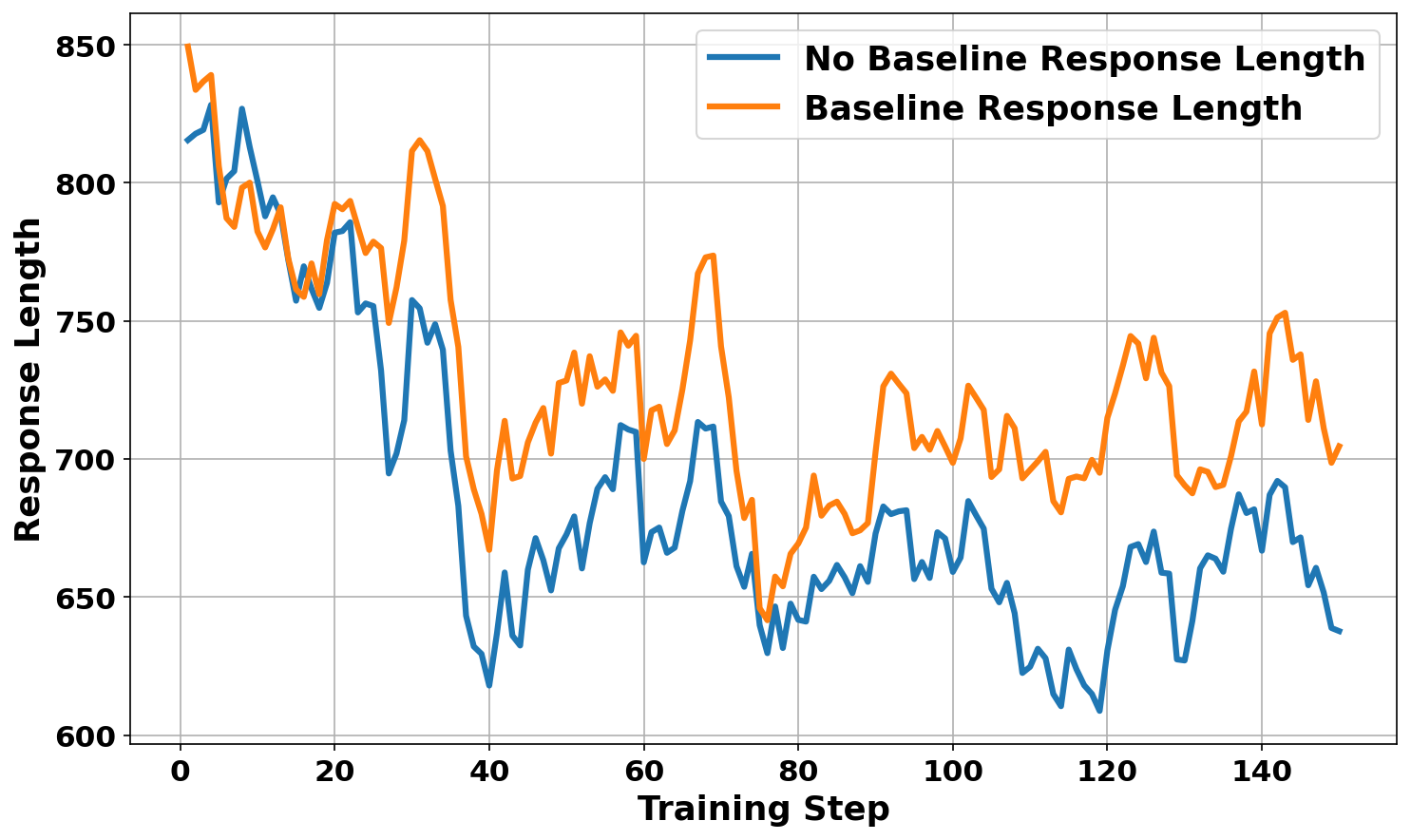}
        \caption{Response Length.}
    \end{subfigure}
    
    \caption{Performance comparison of AdvEvo-MARL training with public baseline for advantage estimation (Baseline) and without using public baseline variant (No Baseline).}
    \label{fig:public baseline ablation}
\end{figure}

In this section, we evaluate the effectiveness of our public baseline for advantage estimation in training. As shown in \cref{fig:public baseline ablation}, the included public baseline leads to more stable and efficient learning dynamics. Specifically, the public-baseline configuration consistently improves both task utility and system safety, as reflected by steadily improved accuracy and controlled ASR during training and evaluation (\cref{fig:ablation asr acc}).
In contrast, the standard training setup exhibits non-stationary behavior and even degraded performance in later stages. 
Moreover, we observe a notable reduction in defender response length under the standard setup, approximately a 13.3\% drop during last 50 training steps, indicating a tendency to produce shorter, less informative outputs. This behavior reflects a defensive overcompensation aimed at minimizing risk, but at the cost of task completeness.
Finally, as shown in the attacker’s reward trajectory, defenders trained with the public baseline are more effective in suppressing the attacker, leading to lower attacker rewards over time, and the defenders also achieve higher rewards via the entire training course. These results provide additional evidence that public-baseline training fosters more robust and generalizable defense policies under adversarial pressure.

\section{Conclusion}
We propose AdvEvo-MARL, a multi-agent safety training framework that enhances the robustness and utility of multi-agent systems through co-evolutionary reinforcement learning. 
By co-training attackers and defenders in a dynamic adversarial environment,
AdvEvo-MARL enables agents to continuously adapt to evolving threats, developing stronger and more generalizable defense capabilities.
To facilitate collaborative learning and stabilize training, we introduce a public baseline mechanism for advantage estimation, where agents within the same role group (e.g., attackers or defenders) share a common baseline calculated from group-level returns. Extensive experiments across representative attack strategies and task benchmarks demonstrate that AdvEvo-MARL substantially improves system safety while preserving and even enhancing task performance. These results highlight AdvEvo-MARL as a promising and unified framework for building safe and capable multi-agent systems.

\clearpage
\section{Ethics Statement}
Our work aims to enhance safety and task utility of multi-agent systems by explicitly addressing their vulnerabilities through co-evolutionary training of attackers and defenders. We adopt a proactive approach that surfaces how current MAS can be compromised and how defense capabilities can be internalized via reinforcement learning. While our methodology involves generating adversarial attacks, these are used solely for the purpose of strengthening defense mechanisms within multi-agent systems. We believe that open, systematic study of such adversarial attacks is critical for the development of safe and resilient AI systems, and ensuring that MAS have broader positive social impacts.

\bibliography{iclr2026_conference}
\bibliographystyle{iclr2026_conference}

\end{document}